\documentclass[11pt]{article}
\usepackage{coling2020}
\usepackage[dvipsnames]{xcolor}
\usepackage{times}
\usepackage{url}
\usepackage{latexsym}

\usepackage{soul}
\usepackage[hidelinks]{hyperref}
\usepackage[utf8]{inputenc}
\usepackage[small]{caption}
\usepackage{graphicx}
\usepackage{amsmath}
\usepackage{amsthm}
\usepackage{booktabs}
\usepackage{algorithm}
\usepackage{algorithmic}
\usepackage{amsfonts}
\usepackage{pgfplots}
\usepackage{subfigure}
\usepackage{multirow}
\usepackage{booktabs}
\usepackage{marvosym}

\colingfinalcopy 

\title{Robust Unsupervised Neural Machine Translation \\with Adversarial Denoising Training}

\author{ Haipeng Sun{$^1$}\thanks{\;\;Most of this work was done when Haipeng Sun was an internship research fellow at NICT.}, Rui Wang{$^2$},  Kehai Chen{$^2$}, Xugang Lu{$^2$}, \\ \textbf{Masao Utiyama{$^2$}, Eiichiro Sumita{$^2$}, and Tiejun Zhao{$^1$}\thanks{\;\;Corresponding author.}} \\
	$^1$Harbin Institute of Technology, Harbin, China \\
	$^2$National Institute of Information and Communications Technology (NICT), Kyoto, Japan \\
	\texttt{sunhaipeng.nlp@gmail.com,} \texttt{tjzhao@hit.edu.cn} \\
	\texttt{\{wangrui,khchen,xugang.lu,mutiyama,eiichiro.sumita\}@nict.go.jp} \\
	\\}

\date{}

\begin{document}
\maketitle
\begin{abstract}
Unsupervised neural machine translation (UNMT) has recently attracted great interest in the machine translation community. The main advantage of the UNMT lies in its easy collection of required large training text sentences while with only a slightly worse performance than supervised neural machine translation which requires expensive annotated translation pairs on some translation tasks. In most studies, the UMNT is trained with clean data without considering its robustness to the noisy data. However, in real-world scenarios, there usually exists noise in the collected input sentences which degrades the performance of the translation system since the UNMT is sensitive to the small perturbations of the input sentences. In this paper, we first time explicitly take the noisy data into consideration to improve the robustness of the UNMT based systems. First of all, we clearly defined two types of noises in training sentences, i.e., word noise and word order noise, and empirically investigate its effect in the UNMT, then we propose adversarial training methods with denoising process in the UNMT. Experimental results on several language pairs show that our proposed methods substantially improved the robustness of the conventional UNMT systems in noisy scenarios.
\end{abstract}
\section{Introduction}
\blfootnote{
    \hspace{-0.65cm}  
    This work is licensed under a Creative Commons 
    Attribution 4.0 International License.
    License details:
    \url{http://creativecommons.org/licenses/by/4.0/}.
}

Recently, unsupervised neural machine translation (UNMT) has attracted great interest in the machine translation community~\cite{DBLP:journals/corr/abs-1710-11041,lample2017unsupervised,P18-1005,lample2018phrase,sun-etal-2019-unsupervised,9043536}.
Typically, UNMT relies solely on monolingual corpora rather than bilingual parallel data in supervised neural machine translation (SNMT) to model translations between the source language and target language and has achieved remarkable results on several translation tasks~\cite{DBLP:journals/corr/abs-1901-07291}. However, previous work only focus on how to build state-of-the-art UNMT systems on the clean data and ignore the robustness of UNMT on the noisy data.
In the real-world scenario, there often exists noises or perturbations in the input sentences, for example, word character misspelling, replacement, or word position misordering, etc. The translation model is sensitive to these perturbations, leading to various errors even the perturbations are small. The existing neural translation system, which lacks of robustness, is difficult to be widely applied to the noisy-data scenario (denoted as noisy scenario in the following sections). Therefore, the  robustness of neural translation system is not only worthy of being studied, but also very essential in the real-world scenarios.

The robustness of SNMT \cite{DBLP:conf/iclr/BelinkovB18,cheng-etal-2018-towards,cheng-etal-2019-robust,DBLP:conf/aclnut/KarpukhinLEG19} has been well-studied. However, most previous work only focus on the effect of the word substitution for translation performance, and ignore the effect of word order for translation performance. Moreover, the noisy robustness of UNMT is much more difficult since the noisy input data may be relieved in some degree by the SNMT due to its supervised check in training. Currently, there is no study considering the noisy robustness of the UNMT. In this paper, we first define two types of noises which cover the noise types mentioned above, i.e., word noise and word order noise. Then we empirically investigate the performance of UNMT in these noisy scenarios. Our empirical results show
that the performance of UNMT model 
decreased substantially, regardless of any noisy scenario. To improve the robustness, we proposed adversarial training methods to alleviate the poor performance in these noisy scenarios. To the best of our knowledge, this paper is the first work to explore the robustness of UNMT. Experimental results on several language pairs show that the proposed strategies substantially outperform conventional UNMT systems in the noisy scenarios.

Our main contributions are summarized as follows:

\begin{itemize}
	\item  We explicitly defined two types of noises, i.e., word noise and word order noise, and empirically investigate the performance of  UNMT in the noisy scenarios.
	
	\item We propose adversarial training methods with denoising process in UNMT training to improve the robustness of the UNMT systems.
	
	\item Our proposed adversarial training methods achieve improvement up to 10 BLEU scores in the noisy scenarios, compared with an UNMT based baseline system.
\end{itemize}


\section{Background of UNMT}
\label{sec:second}
There are four primary components of the state-of-the-art UNMT~\cite{DBLP:journals/corr/abs-1901-07291}: cross-lingual language model pre-training, denoising auto-encoder, back-translation, and sharing latent  representations.	
Consider monolingual data $\{X_i\}$  in language $L_1$ and $\{Y_i\}$ in another language $L_2$.  $|X|$ and $|Y|$ are the number of sentences in monolingual corpora $\{X_i\}$ and $\{Y_i\}$ respectively.
The encoders and decoders of $L_1,L_2$ are trained through denoising and back-translation. 
The objective function $\mathcal{L}_{all}$ of the entire UNMT model would be optimized as:

{\footnotesize
\begin{equation}
	\begin{aligned}
		\mathcal{L}_{all} = \mathcal{L}_{D} + \mathcal{L}_{B},
	\end{aligned}
\end{equation}}%
where $ \mathcal{L}_{D}$ is the objective function for denoising, and $ \mathcal{L}_{B}$ is the objective function for back-translation.

\noindent \textbf{Cross-lingual language model pre-training}: It aims at building a universal cross-lingual encoder that can encode two monolingual sentences into a shared embedding space.
The pre-trained cross-lingual encoder is then used to initialize the UNMT model.

\noindent \textbf{Denoising auto-encoder}: 
In contrast with the normal auto-encoder, denoising auto-encoder~\cite{DBLP:journals/jmlr/VincentLLBM10} could  improve the model learning ability by introducing noise in the form of random token deleting and swapping  in this input sentence.
The denoising auto-encoder, which encodes a noisy version and reconstructs it with the decoder in the same language, acts as a language model during UNMT training.
It is optimized by minimizing the objective function:

{\footnotesize
\begin{equation}
	\begin{aligned}
		\mathcal{L}_{D} &= \sum_{i=1}^{|X|} -\log P_{L_1 \to L_1}(X_i|C(X_i)) + \sum_{i=1}^{|Y|} -\log P_{L_2 \to L_2}(Y_i|C(Y_i)),
	\end{aligned}
\end{equation}}%
where $\{C(X_i)\}$ and $\{C(Y_i)\}$ are noisy sentences. $P_{L_1 \to L_1}$ and $P_{L_2 \to L_2}$ denote the reconstruction probability in the language  $L_1$ and $L_2$, respectively. 

\noindent \textbf{Back-translation}:
It~\cite{P16-1009} is adapted to train a translation system across different languages based on monolingual corpora. The pseudo-parallel sentence pairs $\{(Y_M(X_i),X_i)\}$ and $\{(X_M(Y_i),Y_i)\}$ produced by the model at the previous iteration would be used to   train the new translation model. The UNMT model would be improved through iterative back-translation. Therefore, the back-translation probability would be optimized by minimizing

{\footnotesize
\begin{equation}
	\begin{aligned}
		\mathcal{L}_{B} &= \sum_{i=1}^{|X|} -\log P_{L_2 \to L_1}(X_i|Y_M(X_i)) +\sum_{i=1}^{|Y|} -\log P_{L_1 \to L_2}(Y_i|X_M(Y_i)),
	\end{aligned}
\end{equation}}%
where $P_{L_1 \to L_2}$ and $P_{L_2 \to L_1}$ denote the translation probability across the two languages.

\noindent \textbf{Sharing latent  representations}: 
The same vocabulary is used for both languages. Encoders and decoders are shared for both languages, to help  UNMT model to translate more fluently with synthetic source sentences benefiting from denoising training.

\section{Preliminary Experiments toward Noisy Input to UNMT System}
\label{sec:fourth}
In this section, we first introduce the two primary types of noises in the corpus of UNMT. Then we empirically analyze the effect of these noises on UNMT.

\subsection{Synthetic Noise Generation}
\label{sec:sng}
A few studies of the SNMT robustness~\cite{DBLP:conf/iclr/BelinkovB18,DBLP:conf/aclnut/KarpukhinLEG19} focus on  character-level noise, which affects the spelling of a single word. In this paper, we study the word-level noise, which affects the meaning of a word in one sentence. We refer to this noise as word noise in this paper. Moreover, we study the sentence-level noise, which affects the order of a whole sentence. We refer to this noise as word order noise in this paper.

\noindent \textbf{Word Noise:} We replace every word in the source sentence  by an arbitrary word with a probability $a$.  A larger probability $a$ results in that  more words are replaced by arbitrary words.

\noindent \textbf{Word Order Noise:}
Motivated by the input shuffling strategy~\cite{lample2017unsupervised}, we apply a random permutation $\gamma$ to the source sentence to change the word order of the original sentence, to meet the condition:

{\footnotesize
\begin{equation}
	\begin{aligned}
		\lvert\gamma(i)-i\rvert\leq b, \forall i \in \{1,n\},
	\end{aligned}
\end{equation}}%
where $n$ denotes the length of the source sentence, and $b$ is a hyper-parameter to control the magnitude of the word order adjustment. A larger $b$ results in worse order in the source sentence.

To generate a random permutation verifying the above condition for a sentence of size $n$, we generate a random array $Q$ of size $n$:

{\footnotesize
\begin{equation}
	\begin{aligned}
		Q_i = i + U(0,b),
	\end{aligned}
\end{equation}}%
where  $U$ is the uniform distribution function in the range from $0$ to $b$. Then,  $\gamma$ is defined to be the permutation that sorts the array $Q$. We apply this permutation to adjust word order in order to generate synthetic noise. Note that the order will be changed only when $b > 1$.

\subsection{Noisy Scenario}

As the cross-lingual language model pre-training, which needs large-scale additional monolingual data,  the UNMT system performed comparable with SNMT system which only relies on parallel data. To investigate the performance of UNMT in the noisy scenario,
we empirically choose English (En)--French (Fr) as the language pair to do stimulated experiments.
The detailed experimental settings for UNMT are given in Section~\ref{sec:experiment}. 
The training data is clean and the test data contains synthetic noises described in Section~\ref{sec:sng}. 

\begin{figure*}[ht]
	\centering
	\scalebox{.93}{
	\begin{minipage}[b]{0.48\textwidth}
		\setlength{\abovecaptionskip}{0pt}
		\begin{center}
			\pgfplotsset{height=5.6cm,width=8.5cm,compat=1.14,every axis/.append style={thick},every axis legend/.append style={
at={(1,1)}},legend columns=1}
			\begin{tikzpicture}
			\tikzset{every node}=[font=\small]
			\begin{axis}
			[width=7cm,enlargelimits=0.13, tick align=outside, xticklabels={ $0$, $0.05$,$0.1$, $0.15$, $0.2$,$0.25$},
 axis y line*=left,
xtick={0,1,2,3,4,5},
 ylabel={Translation BLEU score},
 axis x line*=left, 
 ylabel style={align=left},xlabel={$a$ value},font=\small]
			\addplot+ [sharp plot,mark=square*,mark size=1.2pt,mark options={solid,mark color=orange}, color=orange] coordinates
			{ (0,37.25)(1,29.74)(2,25.35)(3,22.92)(4,21.33)(5,19.71) };\label{plot_bb}
			\addlegendentry{\tiny En--Fr}
						\addplot+ [sharp plot,mark=square*,mark size=1.2pt,mark options={solid,mark color=cyan}, color=cyan] coordinates
			{ (0,34.45)(1,26.51)(2,22.32)(3,20.59)(4,18.77)(5,17.57) };\label{plot_aa}
			\addlegendentry{\tiny Fr--En}

\end{axis}
\begin{axis}
[width=7cm,enlargelimits=0.13, tick align=outside,  xticklabels={ $0$, $0.05$,$0.1$, $0.15$, $0.2$,$0.25$},
axis y line*=right,
axis x line=none,
xtick={0,1,2,3,4,5},
 ylabel={Auto-encoder BLEU score},xlabel={$a$ value},font=\small]
 
\addlegendimage{/pgfplots/refstyle=plot_bb}\addlegendentry{\tiny En--Fr} 
\addlegendimage{/pgfplots/refstyle=plot_aa}\addlegendentry{\tiny Fr--En}
		\addplot+ [sharp plot,densely dashed,mark=triangle*,mark size=1.2pt,mark options={solid,mark color=orange}, color=orange] coordinates
			{ (0,94.57)(1,91.31)(2,85.82)(3,80.12)(4,74.59)(5,67.72) };
			\addlegendentry{\tiny En--En}
						\addplot+ [sharp plot,densely dashed,mark=triangle*,mark size=1.2pt,mark options={solid,mark color=cyan}, color=cyan] coordinates
			{ (0,94.65)(1,93.37)(2,88.98)(3,83.91)(4,78.87)(5,73.64) };
			\addlegendentry{\tiny Fr--Fr}
			
			\end{axis}
			\end{tikzpicture}

		\end{center}
	\end{minipage}
	\begin{minipage}[b]{0.48\textwidth}
			\setlength{\abovecaptionskip}{0pt}
		\begin{center}
			\pgfplotsset{height=5.6cm,width=8.5cm,compat=1.14,every axis/.append style={thick},every axis legend/.append style={
at={(1,1)}},legend columns=1}
			\begin{tikzpicture}
			\tikzset{every node}=[font=\small]
			\begin{axis}
			[width=7cm,enlargelimits=0.13, tick align=outside, xticklabels={$0$, $2$,$3$, $5$, $8$,$10$},
 axis y line*=left,
xtick={0,1,2,3,4,5},
 ylabel={Translation BLEU score},
 axis x line*=left,
 ylabel style={align=left},xlabel={$b$ value},font=\small]
			\addplot+ [sharp plot,mark=square*,mark size=1.2pt,mark options={solid,mark color=orange}, color=orange] coordinates
			{ (0,37.25)(1,32.82)(2,26.52)(3,20.3)(4,13.4)(5,11.07) };\label{plot_ba}
			\addlegendentry{\tiny En--Fr}
					\addplot+ [sharp plot,mark=square*,mark size=1.2pt,mark options={solid,mark color=cyan}, color=cyan] coordinates
			{ (0,34.45)(1,29.72)(2,23.20)(3,17.94)(4,12.75)(5,10.45) };\label{plot_ab}
			\addlegendentry{\tiny Fr--En}
\end{axis}
\begin{axis}
[width=7cm,enlargelimits=0.13, tick align=outside,  xticklabels={$0$, $2$,$3$, $5$, $8$,$10$},
axis y line*=right,
axis x line=none,
xtick={0,1,2,3,4,5},
 ylabel={Auto-encoder BLEU score},xlabel={$b$ value},font=\small]
 
\addlegendimage{/pgfplots/refstyle=plot_ba}\addlegendentry{\tiny En--Fr} 
\addlegendimage{/pgfplots/refstyle=plot_ab}\addlegendentry{\tiny Fr--En}
			\addplot+ [sharp plot,densely dashed,mark=triangle*,mark size=1.2pt,mark options={solid,mark color=orange}, color=orange] coordinates
			{ (0,94.57)(1,94.54)(2,92.59)(3,79.55)(4,50.83)(5,39.58) };
			\addlegendentry{\tiny En--En}
						\addplot+ [sharp plot,densely dashed,mark=triangle*,mark size=1.2pt,mark options={solid,mark color=cyan}, color=cyan] coordinates
			{ (0,94.65)(1,95.94)(2,95.45)(3,85.53)(4,59.52)(5,45.66) };
			\addlegendentry{\tiny Fr--Fr}
			
			\end{axis}
			\end{tikzpicture}

		\end{center}
	\end{minipage}}
	\caption{\label{fig:lambda} The UNMT performance as the word noise ($a$ value) and word order noise ($b$ value) increases on the noisy En-Fr newstest2014 set.}
\end{figure*}
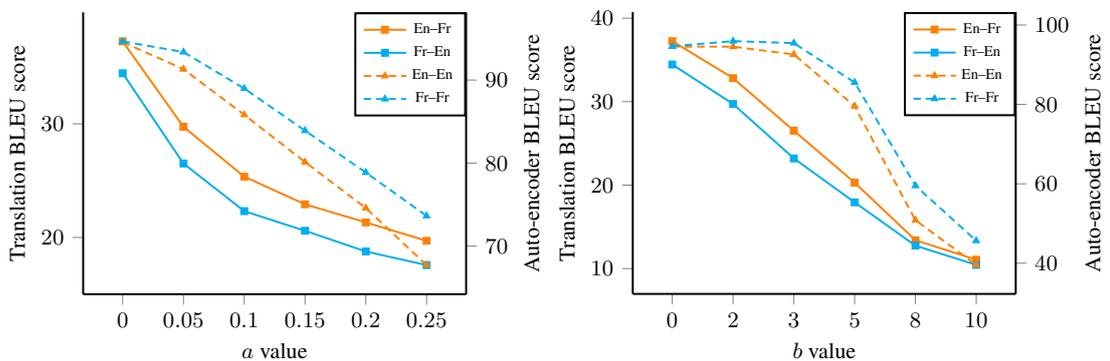	

Figure~\ref{fig:lambda} shows the  BLEU scores of UNMT system with different settings of word noise (left panel) and word order noise (right panel). 
As shown in Figure~\ref{fig:lambda}, we can see that as the ratio of noise in the source language input increased, the performance of both the translation direction (En--Fr and Fr--En) and auto-encoder direction (En--En and Fr--Fr) of UNMT system decreased.  
In addition, when a slight noise ($a\leq0.1$ or $b\leq3$) was added into the input sentence, the translation direction of UNMT still decreased rapidly while a slight downward trend could be maintained in the auto-encoder direction. From these results, we conformed that even a slight input noise could drastically degrade the translation performance of a UNMT system. It is necessary for us to figure out robust solutions to deal with these two types of input noise.

\section{Proposed Methods}
\label{sec:fifth}
\begin{figure*}[thb!]
	\centering
	\includegraphics[width=0.72\textwidth]{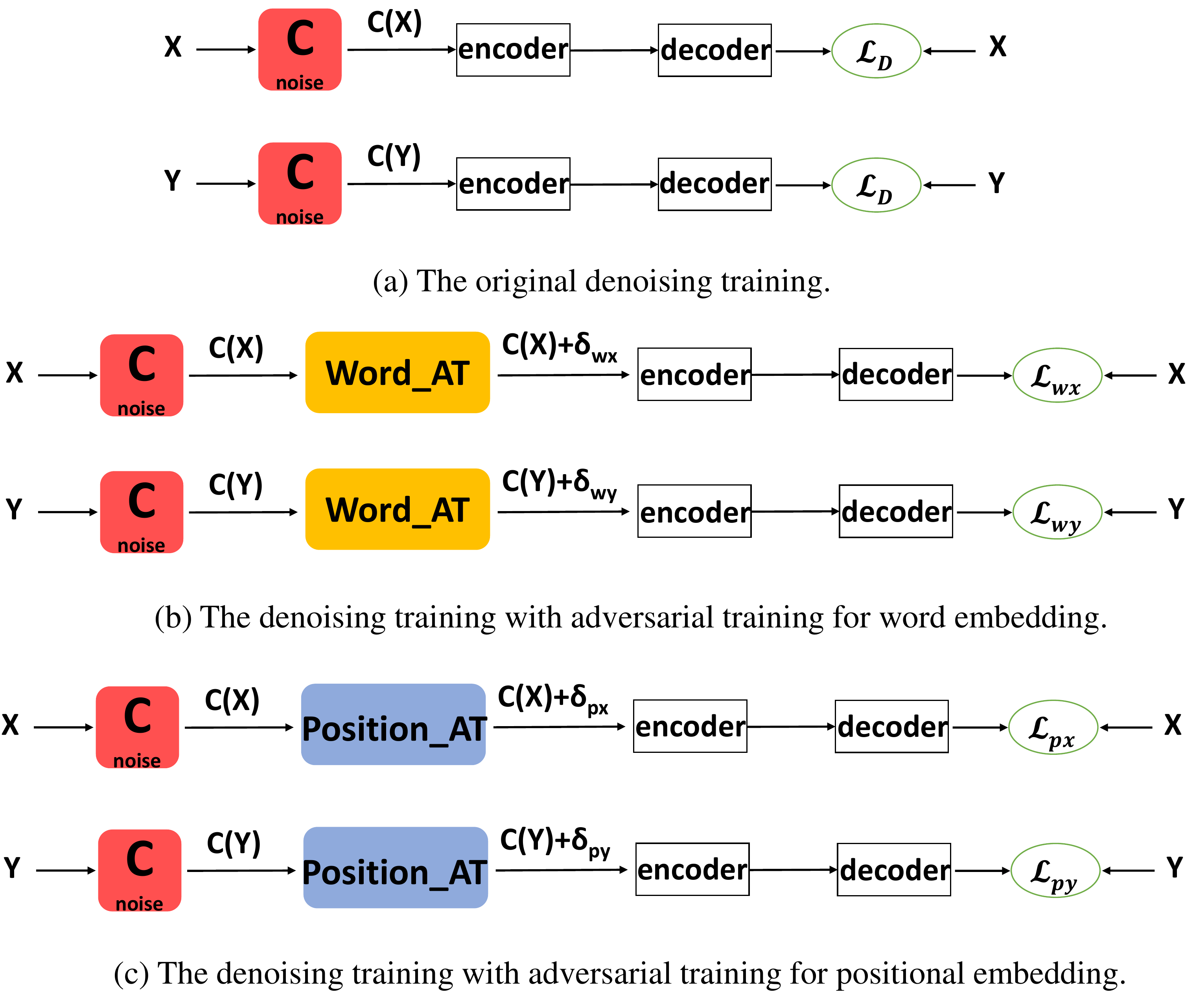}
	\caption{ Illustration of (a) the original denoising process; (b) the denoising process with adversarial training for word embedding (Word\_AT); (c) the denoising process with adversarial training for positional embedding (Position\_AT). \label{fig:architecture}} 
\end{figure*}

Based on the previous empirical findings and analysis, we propose two adversarial training methods during denoising training to improve the robustness of UNMT in the two noisy scenarios as defined in Section~\ref{sec:sng}. The frameworks are illustrated in Figure~\ref{fig:architecture}. The Figure~\ref{fig:architecture} (a) is the original denoising training framework in which an encoder-decoder structure is applied.  Based on this original framework, two adversarial training frameworks are proposed (Figure~\ref{fig:architecture} (b) and (c)) in which the word noise and word-order noise blocks are explicitly modeled.  Although the 
original structure has the denoising ability to replace some wrong words and adjust word order, the denoising ability is implicitly modeled without considering the embedding representation effect which is important in translation. As an adversarial training, our two proposed frameworks explicitly model the two noise effects on the word embedding and position embedding which is expected to improve the translation performance. The details are explained in the following subsections.
\subsection{Adversarial Training for Word Embedding}

The adversarial training strategy~\cite{DBLP:conf/iclr/MiyatoDG17} has been applied to text classification when the input text is noisy. Inspired by this strategy, we apply adversarial training method to the denoising process of UNMT. Regarding to the original denoising process, adversarial perturbation would be added to enhance the learning ability of the denoising auto-encoder. 
Adversarial perturbation is added to the word embedding as a combined word embedding before combining with the positional embedding, compared to the original transformer based architecture. 

As a denoising auto-encoder, for an input of language $L_1$ with the worst case perturbation $\delta_{wx}$ on word embedding, the purpose is to recover the clean version. It is realized by minimizing the reconstruction error defined as: 

{\footnotesize
\begin{equation}
	\begin{aligned}
		\mathcal{L}_{wx} = \sum_{i=1}^{|X|} -\log P_{L_1 \to L_1}(X_i|C(X_i)+\delta_{wx}), \\
		\label{eq:x}
	\end{aligned}
\end{equation}%
\begin{equation}
	\begin{aligned}
		\delta_{wx} = \mathop{\arg\max}_{\ \ \| \delta\|\leq\epsilon }  -\log P_{L_1 \to L_1}(X_i|C(X_i)+\delta),
		\label{eq:atx}
	\end{aligned}
\end{equation}}%
where $\delta$ is a small perturbation in the source side. Actually, it is intractable to calculate the maximization of objective function as shown in Eq. \ref{eq:atx}. Following  \newcite{DBLP:conf/iclr/MiyatoDG17}'s method, the word adversarial perturbation $\delta_{wx}$ is approximated via the gradient of objective function as

{\footnotesize
\begin{equation}
	\begin{aligned}
		\delta_{wx} = \epsilon g_x \  /  \ \| g_x\|_2,\\ \label{eq:px1}
	\end{aligned}
\end{equation}%
\begin{equation}
	\begin{aligned}
		g_x = \bigtriangledown_x -\log P_{L_1 \to L_1}(X_i|C(X_i)),\\  \label{eq:px2}
	\end{aligned}
\end{equation}}%
where $g_x$ denotes the gradient of objective function, calculated by back-propagation algorithm. $\epsilon$ is a hyper-parameter to control the magnitude of adversarial perturbation.

The word adversarial perturbation objective function for language $L_2$ is similarly optimized as

{\footnotesize
\begin{equation}
	\begin{aligned}
		\mathcal{L}_{wy} = \sum_{i=1}^{|Y|} -\log P_{L_2 \to L_2}(Y_i|C(Y_i)+\delta_{wy}), \\
		\label{eq:y}
	\end{aligned}
\end{equation}%
\begin{equation}
	\begin{aligned}
		\delta_{wy} = \epsilon g_y/\ \ \| g_y\|_2,\\
	\end{aligned}
\end{equation}%
\begin{equation}
	\begin{aligned}
		g_y = \bigtriangledown_y -\log P_{L_2 \to L_2}(Y_i|C(Y_i)).
	\end{aligned}
\end{equation}}%

Typically,  to improve UNMT robustness, objective function $\mathcal{L}_{wx}$ and $\mathcal{L}_{wy}$ would be added during the UNMT denoising training process.
The entire UNMT objective function is reformulated as follows:

{\footnotesize
\begin{equation}
	\begin{aligned}
		\mathcal{L}_{all} = \mathcal{L}_{D'} + \mathcal{L}_{B},
	\end{aligned}
\end{equation}%
\begin{equation}
	\begin{aligned}
		\mathcal{L}_{D'} = \mathcal{L}_{D} + \mathcal{L}_{wx} + \mathcal{L}_{wy}.
	\end{aligned}
\end{equation}}%
\subsection{Adversarial Training for Positional Embedding}
Word order is very important in translation. In the state-of-the art NMT framework, the positional embedding has been used to encode order information for the source sentence in the transformer based architecture.
To capture the order robustness, we propose adversarial training method based on the original positional embedding.

The adversarial perturbation is added to the original positional embedding as a new positional embedding before combining with the word embedding, compared to the original transformer architecture.
The positional adversarial perturbation $\delta_{px}$ is then used to penalize the existing positional embedding of the source sentence to generate a new order positional embedding during UNMT denoising training for both languages. 
Similarly as defined in denosing training for word embedding, the objective function for denosing of the positional perturbation is defined as

{\footnotesize
\begin{equation}
	\begin{aligned}
		\mathcal{L}_{px} = \sum_{i=1}^{|X|} -\log P_{L_1 \to L_1}(X_i|C(X_i)+\delta_{px}), \\
		\label{eq:px}
	\end{aligned}
\end{equation}%
\begin{equation}
	\begin{aligned}
		\mathcal{L}_{py} = \sum_{i=1}^{|Y|} -\log P_{L_2 \to L_2}(Y_i|C(Y_i)+\delta_{py}), \\
		\label{eq:y}
	\end{aligned}
\end{equation}}%

To improve UNMT robustness for word order noise, the objective function $\mathcal{L}_{px}$ and $\mathcal{L}_{py}$ would be added during the UNMT denoising training process.
The entire denoising objective function is reformulated as follows:

{\footnotesize
\begin{equation}
	\begin{aligned}
		\mathcal{L}_{D'} = \mathcal{L}_{D} + \mathcal{L}_{px} + \mathcal{L}_{py}.
	\end{aligned}
\end{equation}}%

\subsection{UNMT with Adversarial Training Mechanism}	
Based on our proposed adversarial training methods,  
 we design three UNMT systems with adversarial training: adversarial training for word embedding (Word\_AT), adversarial training for positional embedding (Position\_AT), and the combination of word and positional adversarial training (Both\_AT), all of which enrich robust information via adversarial perturbation.	




\section{Experiments}
\label{sec:experiment}
\subsection{Datasets}
We considered two language pairs to do simulated  experiments on the Fr$\leftrightarrow$En and German(De)$\leftrightarrow$En translation tasks. 
We used 50 million sentences from WMT monolingual news crawl datasets for each language. 
To make our experiments comparable with previous work \cite{DBLP:journals/corr/abs-1901-07291}, we reported results on  newstest2014 for Fr$\leftrightarrow$En and  newstest2016 for De$\leftrightarrow$En.

For preprocessing, we used \texttt{Moses} tokenizer \cite{koehn-etal-2007-moses}\footnote{\url{https://github.com/moses-smt/mosesdecoder}} for all languages.  For cleaning, we only applied the
\texttt{Moses} script \texttt{clean-corpus-n.perl} to remove lines in the monolingual data containing more than 50 tokens.
For BPE~\cite{sennrich2015neural},  we used a shared vocabulary for every language pair with 60K subword tokens based on BPE. 

\subsection{UNMT Settings}
We used a transformer-based \texttt{XLM} toolkit\footnote{\url{https://github.com/facebookresearch/XLM}} and followed settings of \newcite{DBLP:journals/corr/abs-1901-07291}
for UNMT: 6 layers for the encoder and the decoder. 
The dimension of hidden layers was set to 1024. 
The Adam optimizer \cite{kingma2014adam} was used to optimize the model parameters.
The initial learning rate  was 0.0001,  $\beta_1 = 0.9$, and $\beta_2 = 0.98$. 
The cross-lingual language model was used to pretrain the encoder and decoder of the whole UNMT model. We used 8 NVIDIA V100 GPUs with a batch size of 2,000 tokens per GPU for UNMT training.
We used the case-sensitive 4-gram BLEU score computed by \texttt{multi-bleu.perl} script from \texttt{Moses}~\cite{koehn-etal-2007-moses} to evaluate the performance on test sets.
\begin{table*}[t]
	\centering
	\scalebox{.86}{
		\begin{tabular}{cclllll}
			\toprule
			
			Word Noise &Word Order Noise &Method&En-Fr & Fr-En &En-De&De-En \\
			\midrule
			\multirow{4}{*}{No}	&\multirow{4}{*}{No}	&\newcite{lample2017unsupervised} &15.05&14.31&  9.64&13.33\\
			&&\newcite{DBLP:journals/corr/abs-1710-11041} &15.13&15.56&n/a &n/a\\
			&&\newcite{lample2018phrase} &27.60&27.68&20.23&25.19\\
			&	&\newcite{DBLP:journals/corr/abs-1901-07291} &33.40&33.30&26.40&34.30\\
			\midrule
			\multirow{4}{*}{No}&\multirow{4}{*}{No}&UNMT&37.25&34.32&  26.61  &34.24\\
			&&\;\;\;\;\;\;\;\;\;\;+Word\_AT&  37.68     & 34.88    & 27.55  &  34.39 \\
			&&\;\;\;\;\;\;\;\;\;\;+Position\_AT&  37.72     & 34.75    &  27.48 &  34.41 \\
			&&\;\;\;\;\;\;\;\;\;\;+Both\_AT&  37.81     &   34.90  & 27.61  & 34.43   \\
			\midrule
			\multirow{4}{*}{Yes}&\multirow{4}{*}{No}&UNMT&25.35&22.32& 14.40  &26.29\\
			&&\;\;\;\;\;\;\;\;\;\;+Word\_AT&   32.90    &  30.46   & 24.09  & 30.42  \\
			&&\;\;\;\;\;\;\;\;\;\;+Position\_AT&  25.85     &   23.11  & 15.82  & 27.16  \\
			&&\;\;\;\;\;\;\;\;\;\;+Both\_AT&   33.52    &  31.19   & 24.95  &  30.58 \\
			\midrule
			\multirow{4}{*}{No}&\multirow{4}{*}{Yes}&UNMT&26.52&23.20&  15.83  &27.52\\
			&&\;\;\;\;\;\;\;\;\;\;+Word\_AT&   27.48    &  24.02   & 16.33  &  28.48 \\
			&&\;\;\;\;\;\;\;\;\;\;+Position\_AT&  35.44     & 33.29    &  25.72 & 31.86  \\
			&&\;\;\;\;\;\;\;\;\;\;+Both\_AT&36.29&33.78& 25.86   &32.58\\
			\midrule
			\multirow{4}{*}{Yes}&\multirow{4}{*}{Yes}&UNMT&21.78&19.03&  11.35  & 22.89  \\
			&&\;\;\;\;\;\;\;\;\;\;+Word\_AT&    26.02   &    22.49 &16.07&26.14\\
			&&\;\;\;\;\;\;\;\;\;\;+Position\_AT&    25.07   & 22.02    & 15,39  & 25.77 \\
			&&\;\;\;\;\;\;\;\;\;\;+Both\_AT&   32.30    &  30.75   & 22.59  & 29.08  \\
			
			\bottomrule
			
	\end{tabular}}
	\caption{ Performance (BLEU score) of UNMT in  different level of noisy scenarios. $a = 0.1$ for word noise; $b=3$ for word order noise.}
	\label{tab:main_results}
\end{table*}

\subsection{Main Results}

Table \ref{tab:main_results} shows the detailed BLEU scores of all UNMT systems in different level of noisy scenarios on the Fr$\leftrightarrow$En and De$\leftrightarrow$En test sets. Our observations are as follows:

1) Our re-implemented baseline  in this work outperformed the state-of-the-art method~\cite{DBLP:journals/corr/abs-1901-07291}, using clean input on the Fr$\leftrightarrow$En test set, and achieved performance comparable to the original method on the De$\leftrightarrow$En test sets. This indicates that it is a strong UNMT baseline system. 

2) In the scenario that there  is only word noise in the input, our proposed Word\_AT method substantially outperformed the original baseline by approximately 7.4 BLEU scores. Moreover, in the scenario that there  is only word order noise in the input, our proposed Position\_AT method substantially outperformed the original baseline by approximately 8.3 BLEU scores.

3) In the noisy scenario containing the word noise and word order noise, the performance of the original UNMT system decreased drastically. Our proposed Word\_AT and Position\_AT method achieved average improvements of 4 and 3.3 BLEU scores, respectively. Moreover, Our proposed Both\_AT method could further improve UNMT performance, achieving an average improvement of 10 BLEU scores. This demonstrates that our proposed methods  effectively  alleviate the noisy input issue.

4) Although our designed adversarial training frameworks tries to remove the effect of noise with explicit inserting word noise and word order noise processing blocks, the frameworks could improve the UNMT performance even in clean scenarios, achieving an average improvement of 0.6 BLEU scores on all clean test sets. This suggests that our proposed methods may facilitate the training efficiency of the network.  

\subsection{Analysis}	
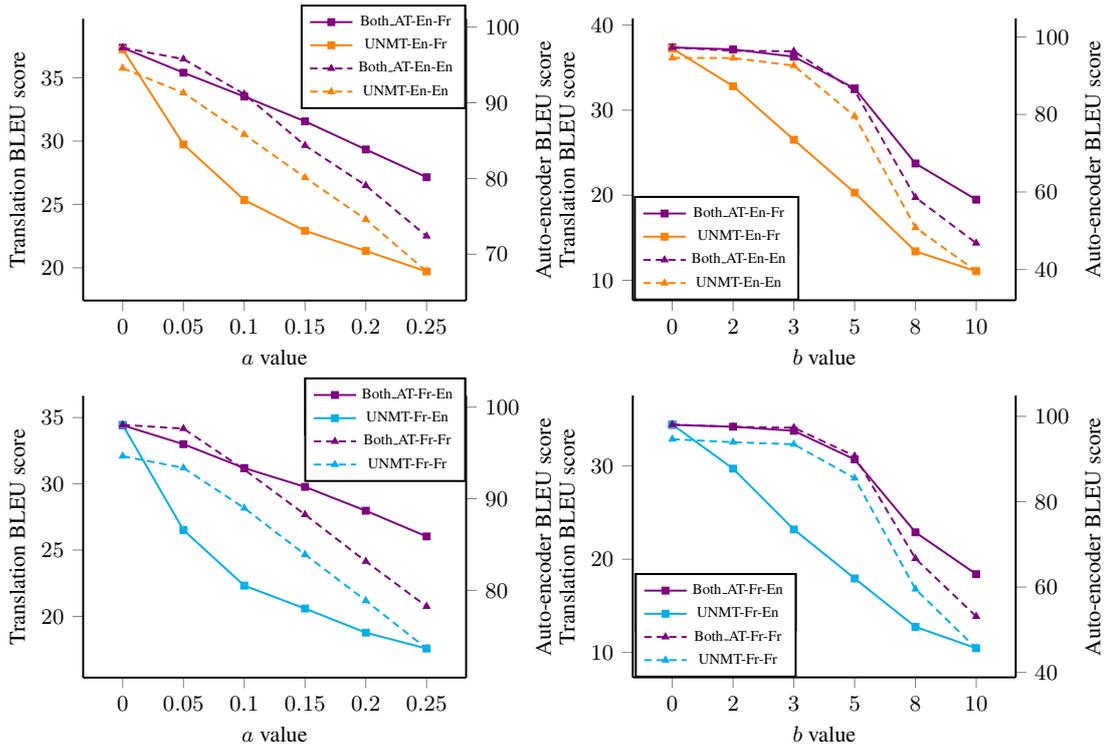
\begin{figure}[ht]
	\centering	\scalebox{.93}{
	\begin{minipage}[b]{0.48\linewidth}
		\setlength{\abovecaptionskip}{0pt}
		\begin{center}
			\pgfplotsset{height=5.6cm,width=8.5cm,compat=1.14,every axis/.append style={thick},legend columns=1,every axis legend/.append style={
					at={(1,1.05)}}}
			\begin{tikzpicture}
			\tikzset{every node}=[font=\small]
			\begin{axis}
			[width=7cm,enlargelimits=0.13, tick align=outside, xticklabels={ $0$, $0.05$,$0.1$, $0.15$, $0.2$,$0.25$},
 axis y line*=left,
xtick={0,1,2,3,4,5},
 ylabel={Translation BLEU score},
 axis x line*=left,
 ylabel style={align=left},xlabel={$a$ value},font=\small]
			
			\addplot+ [sharp plot,mark=square*,mark size=1.2pt,mark options={solid,mark color=violet}, color=violet] coordinates
			{ (0,37.37)(1,35.4)(2,33.52)(3,31.56)(4,29.34)(5,27.15) };\label{plot_b}
			\addlegendentry{\tiny Both\_AT-En-Fr}
			\addplot+ [sharp plot,mark=square*,mark size=1.2pt,mark options={solid,mark color=orange}, color=orange] coordinates
			{ (0,37.25)(1,29.74)(2,25.35)(3,22.92)(4,21.33)(5,19.71) };\label{plot_a}
			\addlegendentry{\tiny UNMT-En-Fr}
\end{axis}
\begin{axis}
[width=7cm,enlargelimits=0.13, tick align=outside,  xticklabels={ $0$, $0.05$,$0.1$, $0.15$, $0.2$,$0.25$},
axis y line*=right,
axis x line=none,
xtick={0,1,2,3,4,5},
 ylabel={Auto-encoder BLEU score},xlabel={$a$ value},font=\small]
 
\addlegendimage{/pgfplots/refstyle=plot_b}\addlegendentry{\tiny Both\_AT-En-Fr} 
\addlegendimage{/pgfplots/refstyle=plot_a}\addlegendentry{\tiny UNMT-En-Fr}			
			
			\addplot+ [sharp plot,densely dashed,mark=triangle*,mark size=1.2pt,mark options={solid,mark color=violet}, color=violet] coordinates
			{ (0,97.27)(1,95.77)(2,91.15)(3,84.36)(4,79.07)(5,72.39) };
			\addlegendentry{\tiny Both\_AT-En-En}
			\addplot+ [sharp plot,densely dashed,mark=triangle*,mark size=1.2pt,mark options={solid,mark color=orange}, color=orange] coordinates
			{ (0,94.57)(1,91.31)(2,85.82)(3,80.12)(4,74.59)(5,67.72) };
			\addlegendentry{\tiny UNMT-En-En}
			
			\end{axis}
			\end{tikzpicture}
		\end{center}
	\end{minipage}
	\begin{minipage}[b]{0.48\linewidth}
		\setlength{\abovecaptionskip}{0pt}
		\begin{center}
			\pgfplotsset{height=5.6cm,width=8.5cm,compat=1.14,every axis/.append style={thick},legend columns=1,every axis legend/.append style={
					at={(0.435,0.37)}}}
			\begin{tikzpicture}
			\tikzset{every node}=[font=\small]
			\begin{axis}
			[width=7cm,enlargelimits=0.13, tick align=outside, xticklabels={ $0$, $2$,$3$, $5$, $8$,$10$},
	axis y line*=left,
xtick={0,1,2,3,4,5},
 ylabel={Translation BLEU score},
 axis x line*=left,
 ylabel style={align=left},xlabel={$b$ value},font=\small]
			
			\addplot+ [sharp plot,mark=square*,mark size=1.2pt,mark options={solid,mark color=violet}, color=violet] coordinates
			{ (0,37.37)(1,37.15)(2,36.29)(3,32.56)(4,23.73)(5,19.48) };\label{plot_d}
			\addlegendentry{\tiny Both\_AT-En-Fr}
			\addplot+ [sharp plot,mark=square*,mark size=1.2pt,mark options={solid,mark color=orange}, color=orange] coordinates
			{ (0,37.25)(1,32.82)(2,26.52)(3,20.3)(4,13.4)(5,11.07) };\label{plot_c}
			\addlegendentry{\tiny UNMT-En-Fr}
\end{axis}
\begin{axis}
[width=7cm,enlargelimits=0.13, tick align=outside,  xticklabels={ $0$, $2$,$3$, $5$, $8$,$10$},
axis y line*=right,
axis x line=none,
xtick={0,1,2,3,4,5},
 ylabel={Auto-encoder BLEU score},xlabel={$b$ value},font=\small]	
 \addlegendimage{/pgfplots/refstyle=plot_d}\addlegendentry{\tiny Both\_AT-En-Fr} 
\addlegendimage{/pgfplots/refstyle=plot_c}\addlegendentry{\tiny UNMT-En-Fr}
				\addplot+ [sharp plot,densely dashed,mark=triangle*,mark size=1.2pt,mark options={solid,mark color=violet}, color=violet] coordinates
			{ (0,97.27)(1,96.41)(2,96.30)(3,86.14)(4,58.59)(5,46.81) };
			\addlegendentry{\tiny Both\_AT-En-En}
			\addplot+ [sharp plot,densely dashed,mark=triangle*,mark size=1.2pt,mark options={solid,mark color=orange}, color=orange] coordinates
			{ (0,94.57)(1,94.54)(2,92.59)(3,79.55)(4,50.83)(5,39.58) };
			\addlegendentry{\tiny UNMT-En-En}
			\end{axis}
			\end{tikzpicture}
		\end{center}
	\end{minipage}} \scalebox{.93}{
		\begin{minipage}[b]{0.48\linewidth}
		\setlength{\abovecaptionskip}{0pt}
		\begin{center}
			\pgfplotsset{height=5.6cm,width=8.5cm,compat=1.14,every axis/.append style={thick},legend columns=1,every axis legend/.append style={
					at={(1,1.065)}}}
			\begin{tikzpicture}
			\tikzset{every node}=[font=\small]
			\begin{axis}
			[width=7cm,enlargelimits=0.13, tick align=outside, xticklabels={ $0$, $0.05$,$0.1$, $0.15$, $0.2$,$0.25$},
 axis y line*=left,
xtick={0,1,2,3,4,5},
 ylabel={Translation BLEU score},
 axis x line*=left,
 ylabel style={align=left},xlabel={$a$ value},font=\small]
						\addplot+ [sharp plot,mark=square*,mark size=1.2pt,mark options={solid,mark color=violet}, color=violet] coordinates
			{ (0,34.42)(1,32.99)(2,31.19)(3,29.77)(4,27.97)(5,26.03) };\label{plot_f}
			\addlegendentry{\tiny Both\_AT-Fr-En}
			\addplot+ [sharp plot,mark=square*,mark size=1.2pt,mark options={solid,mark color=cyan}, color=cyan] coordinates
			{ (0,34.45)(1,26.51)(2,22.32)(3,20.59)(4,18.77)(5,17.57) };\label{plot_e}
			\addlegendentry{\tiny UNMT-Fr-En}
			\end{axis}
\begin{axis}
[width=7cm,enlargelimits=0.13, tick align=outside,  xticklabels={ $0$, $0.05$,$0.1$, $0.15$, $0.2$,$0.25$},
axis y line*=right,
axis x line=none,
xtick={0,1,2,3,4,5},
 ylabel={Auto-encoder BLEU score},xlabel={$a$ value},font=\small]
 
\addlegendimage{/pgfplots/refstyle=plot_f}\addlegendentry{\tiny Both\_AT-Fr-En} 
\addlegendimage{/pgfplots/refstyle=plot_e}\addlegendentry{\tiny UNMT-Fr-En}
			\addplot+ [sharp plot,densely dashed,mark=triangle*,mark size=1.2pt,mark options={solid,mark color=violet}, color=violet] coordinates
			{ (0,98.07)(1,97.66)(2,93.19)(3,88.27)(4,83.16)(5,78.24) };
			\addlegendentry{\tiny Both\_AT-Fr-Fr}
			\addplot+ [sharp plot,densely dashed,mark=triangle*,mark size=1.2pt,mark options={solid,mark color=cyan}, color=cyan] coordinates
			{ (0,94.65)(1,93.37)(2,88.98)(3,83.91)(4,78.87)(5,73.64) };
			\addlegendentry{\tiny UNMT-Fr-Fr}
			
			\end{axis}
			\end{tikzpicture}
		\end{center}
	\end{minipage}
	\begin{minipage}[b]{0.48\linewidth}
		\setlength{\abovecaptionskip}{0pt}
		\begin{center}
			\pgfplotsset{height=5.6cm,width=8.5cm,compat=1.14,every axis/.append style={thick},legend columns=1,every axis legend/.append style={
					at={(0.428,0.37)}}}
			\begin{tikzpicture}
			\tikzset{every node}=[font=\small]
			\begin{axis}
			[width=7cm,enlargelimits=0.13, tick align=outside, xticklabels={ $0$, $2$,$3$, $5$, $8$,$10$},
 axis y line*=left,
xtick={0,1,2,3,4,5},
 ylabel={Translation BLEU score},
 axis x line*=left,
 ylabel style={align=left},xlabel={$b$ value},font=\small]
						\addplot+ [sharp plot,mark=square*,mark size=1.2pt,mark options={solid,mark color=violet}, color=violet] coordinates
			{ (0,34.42)(1,34.22)(2,33.78)(3,30.71)(4,22.89)(5,18.40) };\label{plot_h}
			\addlegendentry{\tiny Both\_AT-Fr-En}
			\addplot+ [sharp plot,mark=square*,mark size=1.2pt,mark options={solid,mark color=cyan}, color=cyan] coordinates
			{ (0,34.45)(1,29.72)(2,23.20)(3,17.94)(4,12.75)(5,10.45) };\label{plot_g}
			\addlegendentry{\tiny UNMT-Fr-En}
\end{axis}
\begin{axis}
[width=7cm,enlargelimits=0.13, tick align=outside,  xticklabels={ $0$, $2$,$3$, $5$, $8$,$10$},
axis y line*=right,
axis x line=none,
xtick={0,1,2,3,4,5},
 ylabel={Auto-encoder BLEU score},xlabel={$b$ value},font=\small]
 
\addlegendimage{/pgfplots/refstyle=plot_h}\addlegendentry{\tiny Both\_AT-Fr-En} 
\addlegendimage{/pgfplots/refstyle=plot_g}\addlegendentry{\tiny UNMT-Fr-En}		
			\addplot+ [sharp plot,densely dashed,mark=triangle*,mark size=1.2pt,mark options={solid,mark color=violet}, color=violet] coordinates
			{ (0,98.07)(1,97.55)(2,97.33)(3,90.70)(4,66.68)(5,53.12) };
			\addlegendentry{\tiny Both\_AT-Fr-Fr}
			\addplot+ [sharp plot,densely dashed,mark=triangle*,mark size=1.2pt,mark options={solid,mark color=cyan}, color=cyan] coordinates
			{ (0,94.65)(1,93.94)(2,93.45)(3,85.53)(4,59.52)(5,45.66) };
			\addlegendentry{\tiny UNMT-Fr-Fr}			
			\end{axis}
			\end{tikzpicture}
		\end{center}
		
	\end{minipage}}

	\caption{\label{fig:noise} The performance of baseline system and UNMT with Both\_AT mechanism as the word noise ($a$ value) and word order noise ($b$ value) increases on the noisy En-Fr newstest2014 set.}
\end{figure}

We empirically investigated the performance of UNMT with Both\_AT framework, using the noisy input with  different level of word noise and word order noise, respectively. Figure \ref{fig:noise} shows the trend measured in BLEU score of UNMT baseline system (orange line) and UNMT system with our proposed Both\_AT framework (purple line) for the translation direction (solid line) and auto-encoder direction (dashed line).
 As shown in Figure \ref{fig:noise}, we can see that our proposed adversarial training method performed significantly better than the UNMT baseline system in the noisy scenario, regardless of any direction. 
 In particular, as the ratios of word noise and word order noise increase, the performance gap between the baseline UNMT system and our proposed system is increased.
 These demonstrate that our proposed adversarial training method is robust and effective.

\begin{table}[h]
	\centering
	\scalebox{.88}{
		\begin{tabular}{ccc|ccc}
			\toprule
$a$ value& UNMT & +Both\_AT & $b$ value & UNMT & +Both\_AT  \\
			
			\midrule
		0&100.00&100.00&0&100.00&100.00\\
	0.05& 52.61&73.79&2&60.28&78.22\\
	0.1&41.09&66.45&3&43.43&74.07\\
	0.15&35.40&60.40&5&31.56&62.99\\
	0.2&32.10&54.42&8&20.15&43.02\\
	0.25&28.48&48.41&10&16.18&33.75\\
			
			\bottomrule
			
	\end{tabular}}
	\caption{Average similarity (BLEU score) across the translation generated by the clean input and noisy input on the En-Fr newstest2014 set. }
	\label{tab:similarity}
\end{table}	

In order to get a more complete picture of the robustness of our proposed adversarial training methods, we empirically evaluated the similarity across the translation generated by the clean input and noisy input with different level of noise on the En-Fr newstest2014 set as shown in Table ~\ref{tab:similarity}. As it can be seen, our proposed Both\_AT mechanism produced substantially more similar translations generated by the noisy input and clean input, compared with the UNMT baseline. As Table ~\ref{tab:similarity} reports, the similarity of our proposed Both\_AT mechanism across the translation generated by the clean input and noisy input with different level of noise was 22.76 BLEU scores more in average for the word noise scenario, and 24.09 BLEU scores more in average for the word order noise scenario, compared with the UNMT system. These further demonstrate that our proposed  Both\_AT mechanism is robust and can effectively alleviate the impact of two types of noise on translation performance.

\subsection{Evaluation on MTNT dataset}
To better assess the effectiveness of our proposed adversarial training methods, we investigated  the performance of UNMT with Both\_AT framework on the  MTNT dataset, which is a noisy dataset proposed by \newcite{DBLP:conf/emnlp/MichelN18}. The detailed statistics of MTNT data set is presented as shown in Table \ref{tab:mtnt}. To make our experiments comparable with previous work \cite{DBLP:conf/emnlp/MichelN18,DBLP:conf/wmt/ZhouZZAN19}, we used the same MTNT parallel training data to fine-tune our proposed +Both\_AT system and used \texttt{sacreBLEU} \cite{DBLP:conf/wmt/Post18} to evaluate the translation performance.

\begin{minipage}{\textwidth}
 \begin{minipage}[t]{0.45\textwidth}
  \centering
     \makeatletter\def\@captype{table}\makeatother
       \begin{tabular}{ccc} 
			\toprule
Corpus& en-fr & fr-en  \\
			
			\midrule
Training set& 36,058   & 19,161 \\
Valid set&  852   &  886  \\
Test set& 1,020   & 1,022    \\
			\bottomrule
	\end{tabular}
	\caption{Statistics of MTNT data set.}
	\label{tab:mtnt}
  \end{minipage}
  \begin{minipage}[t]{0.45\textwidth}
   \centering
        \makeatletter\def\@captype{table}\makeatother
        \scalebox{.85}{
         \begin{tabular}{ccc}        
          	\toprule
Methods& en-fr & fr-en  \\
			
			\midrule
		\newcite{DBLP:conf/emnlp/MichelN18}& 21.77 &23.27  \\
		\;\;\;\;\;\;\;\;\;\;+Fine-tuning&29.73&30.29  \\
		\newcite{DBLP:conf/wmt/ZhouZZAN19}& n/a & 24.50 \\
		\;\;\;\;\;\;\;\;\;\;+Fine-tuning& n/a    & 31.70     \\
		\midrule	
		+Both\_AT  &  31.60 & 33.80\\
		\;\;\;\;\;\;\;\;\;\;+Fine-tuning& 39.00  & 41.30 \\
			\bottomrule
	  \end{tabular}}
	  \caption{BLEU score on the En-Fr MTNT test set. }
	  \label{tab:mtntbleu}
   \end{minipage}
\end{minipage}

As shown in Table \ref{tab:mtntbleu}, Our proposed +Both\_AT system significantly outperformed the previous work\cite{DBLP:conf/emnlp/MichelN18,DBLP:conf/wmt/ZhouZZAN19} by approximately 10 BLEU scores. The performance of our proposed system without fine-tuning was even better than that of these previous work with fine-tuning.
After fine-tuning, our proposed system achieved promising performance, obtaining 39.00 BLEU scores on the en-fr testset and 41.30 BLEU scores on the fr-en testset. These further demonstrate that our proposed +Both\_AT system is a robust system.
\subsection{Case Study}
Moreover, we analyze translation examples to further analyze the effectiveness of our proposed adversarial training method in the noisy scenario.
Table \ref{tab:case} shows two translation examples, which were generated by UNMT baseline system and +Both\_AT system, using clean and noisy input on the Fr-En dataset, respectively. For the first example, +Both\_AT  could adjust the wrong word order to the natural English word order. For the second example, +Both\_AT  could be better at mitigating the impact of missing words and noisy words. These examples indicate that  our proposed +Both\_AT system could be widely applied to  the  real-world noisy scenario.

\begin{table*}[htb]
	\centering
	\scalebox{.95}{
		\begin{tabular}{ll}
			\toprule
			Clean Input& On est très excités, {\color{red}mais très} tendus aussi.\\
			
			Noisy Input& On est très excités, {\color{red}très mais} tendus aussi.\\
			
			Reference& We were very excited, but {\color{PineGreen}also} very tense.\\
			
			\midrule
			Baseline on Clean Input& We're very excited, but very tense {\color{PineGreen}too}.\\
			+Both\_AT on Clean Input& We're very excited, but {\color{PineGreen}also} very tense.\\

			\midrule
			Baseline on Noisy Input&We're very excited, very tense {\color{red}but also}.\\
			
			+Both\_AT on Noisy Input&We're very excited, but {\color{PineGreen}also} very tense.\\	
			\midrule
			\midrule
			Clean Input& {\color{red}Avec} la crise de l' euro, le Projet Europe est officiellement {\color{CadetBlue}mort}.\\
			
			Noisy Input& \;\;\;\;\;\;\;\;la crise de l' euro, le Projet Europe est officiellement {\color{CadetBlue}décès}.\\
			
			Reference& {\color{red}With} the euro crisis, Project Europe is officially {\color{CadetBlue}dead}.\\
			
			\midrule
			Baseline on Clean Input& {\color{red}With} the euro crisis, the Europe Project is officially {\color{CadetBlue}dead}.\\
			
			+Both\_AT on Clean Input& {\color{red}With} the euro crisis, Project Europe is officially {\color{CadetBlue}dead}.\\
			
			\midrule
			
			Baseline on Noisy Input&{\color{red}Amid} the euro crisis, the Europe Project is officially {\color{CadetBlue}a death}.\\
			
			+Both\_AT on Noisy Input&{\color{red}With} the euro crisis, Project Europe is officially {\color{CadetBlue}at its death}.\\					
			
			\bottomrule
			
	\end{tabular}}
	\caption{Comparison of translation results of baseline and +Both\_AT system for clean and noisy input. }
	\label{tab:case}
\end{table*}	

\section{Related Work}
\label{sec:seventh}
Recently, UNMT \cite{DBLP:journals/corr/abs-1710-11041,lample2017unsupervised,P18-1005,lample2018phrase,sun-etal-2019-unsupervised} that relies solely on monolingual corpora in each language via bilingual word embedding initialization, denoising auto-encoder, back-translation and sharing latent representations.
More recently, \newcite{DBLP:journals/corr/abs-1901-07291}  and  \newcite{song2019mass} introduced the pretrained cross-lingual language model to achieve  state-of-the-art UNMT performance. \newcite{sun-etal-2020-knowledge} extended UNMT to the multilingual UNMT training on a large scale of European languages. \newcite{marie-etal-2019-nicts} won the first
place in the unsupervised translation task of WMT19 by combining UNMT and unsupervised statistical machine translation.
However, previous work only focuses  on  how  to  build  state-of-the-art  UNMT systems and ignore the robustness of UNMT on the noisy data. In this paper, we propose adversarial training methods with denoising process in UNMT training to improve the robustness of the UNMT systems. Moreover, our proposed methods could improve the UNMT performance even in clean scenarios.

Actually, \newcite{DBLP:conf/iclr/BelinkovB18} pointed out that synthetic and natural noise both influenced the translation performance.
\newcite{DBLP:conf/iclr/BelinkovB18},
\newcite{ebrahimi-etal-2018-hotflip}, and \newcite{DBLP:conf/aclnut/KarpukhinLEG19} designed character-level noise, which affects the spelling of a single word, to improve the  model robustness. Meanwhile, both textual and phonetic embeddings were used to improve the robustness of SNMT to homophone noises~\cite{liu-etal-2019-robust}. Adversarial examples, generated by gradient-based method, attacked the translation model to improve the robustness of SNMT ~\cite{cheng-etal-2019-robust}. In  contrast  with this work, we applied adversarial perturbation to the denoising training of UNMT , instead of translation training, to enhance the learning ability of UNMT model.

Adversarial training method, first proposed in the computer vision~\cite{DBLP:journals/corr/GoodfellowSS14,DBLP:conf/cvpr/Moosavi-Dezfooli16}, was applied to several natural language processing tasks~\cite{DBLP:conf/iclr/MiyatoDG17,jia-liang-2017-adversarial,DBLP:conf/iclr/BelinkovB18,ebrahimi-etal-2018-hotflip}.
\section{Conclusion}
\label{sec:eighth}

As a data driven machine learning method, neural translation model is sensitive to small data perturbations which may result in poor performance in noisy scenarios. In this paper, we focused on the noisy robust problem in unsupervised machine translation. We first explicitly defined two types of noise which are frequently appeared in real applications. Then we proposed denoising process on word embedding and word positional embedding based adversarial training in the UNMT framework. Experimental results confirmed that the proposed adversarial training significantly improved the robustness of the UNMT systems in the noisy scenarios. Moreover, even for clean scenario, our proposed framework could slightly improve the performance. From the results, we can conclude that our proposed framework could improve the noisy robustness of the UNMT without sacrificing the performance for clean condition. 

In this study, our proposed adversarial training was implemented with the transformer based neural architectures for UNMT. In the future, we will examine the proposed method in other natural language processing tasks and integrate other machine learning methods for the robustness of the UNMT systems.

\section*{Acknowledgments}
Tiejun Zhao was partially supported by National Key Research and Development Program of China via grant 2017YFB1002102.

\bibliographystyle{coling}
\bibliography{coling2020}

\end{document}